\documentclass[10pt,twocolumn,letterpaper]{article}

\usepackage{wacv}
\usepackage{times}
\usepackage{epsfig}
\usepackage{graphicx}
\usepackage{amsmath}
\usepackage{amssymb}
\usepackage{subcaption}
\usepackage{enumitem}

\usepackage{subcaption, multirow, booktabs, url, flushend}
\usepackage[table]{xcolor}

\newcolumntype{L}[1]{>{\raggedright\let\newline\\\arraybackslash\hspace{0pt}}m{#1}}
\newcolumntype{C}[1]{>{\centering\let\newline\\\arraybackslash\hspace{0pt}}m{#1}}
\newcolumntype{R}[1]{>{\raggedleft\let\newline\\\arraybackslash\hspace{0pt}}m{#1}}

\wacvfinalcopy % *** TODO: Uncomment this line for the final submission

\def\wacvPaperID{440}

\ifwacvfinal\fi
\begin{document}

%%%%%%%%% TITLE
\title{Iris Presentation Attack Detection Based on Photometric Stereo Features\thanks{Patent Pending. Paper accepted for WACV 2019, Hawaii, USA}}

% Authors at the same institution
%\author{First Author \hspace{2cm} Second Author \\
%Institution1\\
%{\tt\small firstauthor@i1.org}
%}
% Authors at different institutions
\author{Adam Czajka$^1$ \hspace{1cm} Zhaoyuan Fang$^2$ \hspace{1cm} Kevin W. Bowyer$^1$\\
$^1$Department of Computer Science and Engineering, $^2$Department of Electrical Engineering\\University of Notre Dame\\
{{\tt\small \{aczajka,zfang,kwb\}@nd.edu}}\\
%{\small $^*$Patent Pending}
}

\maketitle
\ifwacvfinal\thispagestyle{empty}\fi

%%%%%%%%% ABSTRACT
\begin{abstract}
We propose a new iris presentation attack detection method using three-dimensional features of an observed iris region estimated by photometric stereo. Our implementation uses a pair of iris images acquired by a common commercial iris sensor (LG 4000). No hardware modifications of any kind are required. Our approach should be applicable to any iris sensor that can illuminate the eye from two different directions. Each iris image in the pair is captured under near-infrared illumination at a different angle relative to the eye. Photometric stereo is used to estimate surface normal vectors in the non-occluded portions of the iris region. The variability of the normal vectors is used as the presentation attack detection score. This score is larger for a texture that is irregularly opaque and printed on a convex contact lens, and is smaller for an authentic iris texture. Thus the problem is formulated as binary classification into (a) an eye wearing textured contact lens and (b) the texture of an actual iris surface (possibly seen through a clear contact lens). Experiments were carried out on a database of approx. 2,900 iris image pairs acquired from approx. 100 subjects. Our method was able to correctly classify over 95\% of samples when tested on contact lens brands unseen in training, and over 98\% of samples when the contact lens brand was seen during training. The source codes of the method are made available to other researchers.
\end{abstract}

%-------------------------------------------------------------------------
\section{Introduction}
\label{sec:Introduction}

Presentation attacks are those presentations to the biometric sensors which aim at either impersonating someone else or concealing the attacker's identity. Such attacks may be carried out in various ways, for instance by presenting artificial objects (\eg gummy finger, face silicon mask, prosthetic eye), non-conformant use of a biometric system (\eg squinting, rotating the finger, changing the face expression), or even presenting cadavers. Iris recognition, considered to be one of the most accurate biometric methods, is relatively easy to spoof and results of the recent LivDet-Iris 2017 competition suggest that both textured contact lens and paper iris printouts are still challenging for state-of-the-art methods \cite{Yambay_IJCB_2014}. The inability to detect textured contact lenses is particularly disturbing, since such lenses may be worn for cosmetic purposes, and thus the attacker may easily conceal the true intention of presenting an eye covered by a textured lens. Wearing a textured contact lens significantly lowers the chances of getting a correct match between the observed eye and the reference template stored in a database \cite{Baker_CVIU_2010,Kohli_ICB_2013,Gupta_ICPR_2014,Yadav_TIFS_2014}.

This is why {\it presentation attack detection} (PAD) is important in biometrics, and automatic recognition of textured contact lenses is crucial for equipment installed in uncontrolled, real-world scenarios, such as a border control. This paper proposes a novel iris PAD method that, different from any previous work known to us, employs photometric stereo to make a distinction between an approximately flat authentic iris, and an irregular and convex shape of the textured contact lens. The practical motivation behind proposing this technique is that the hardware of common commercial sensors is sufficient for implementing this method. Namely, only two near-infrared illuminators\footnote{Iris sensors with illuminators at two locations relative to the camera include the LG 4000, Iris Guard AD 100, Panasonic ET 300 and others. The OKI Iris Pass M even had illuminators in four locations.  Illuminators at two or more locations is common, apparently with the motivation to select images to minimize specular highlights on the iris.}, placed at different locations relative to the camera lens, are enough to estimate the normal vectors for the observed surface. These normal vectors are more consistent for a flat iris than for irregular textured contact lens. The estimated variability of the calculated normals serves as the PAD score, which is smaller for authentic eyes than for eyes wearing textured contact lenses. Consequently, the important {\bf practical advantage} of the proposed method is that it can be seamlessly implemented in present sensors that generate ISO/IEC 19794-6-complaint iris images, without any hardware updates. The source codes of the method are made available along with this paper for non-commercial purposes\footnote{\url{https://github.com/CVRL/PhotometricStereoIrisPAD}}.

%-------------------------------------------------------------------------
\section{Uniqueness Over Related Work}
\label{sec:Related}

Iris presentation attack detection has received considerable attention, especially in recent years. A recent survey by Czajka and Bowyer provides detailed summary of the research to date in iris PAD \cite{Czajka_CSUR_2018}. Below we discuss the iris PAD approaches based on three-dimensional features that are most closely related to the proposed method. In each case we explain how our new approach is different and improves on the known technique.

One approach to distinguish non-eye, flat objects (printed irises) from a living eye is to detect corneal reflections of infrared light illuminated in random sequences, and was envisioned by Daugman \cite{Daugman_WMIP_2003}. The cornea has a spherical shape and, due to its moistness, generates specular reflections at the locations possible to be estimated when the positions of the illuminators are given. Lambertian flat surfaces will not generate specular highlights and thus can be classified as non-eye objects. An algorithm following this concept was patented by Min and Chae \cite{GiMin_USpatent_2004}, and independently proposed later by Pacut and Czajka \cite{Pacut_ICCST_2006}. They used two supplementary sources of near-infrared light placed equidistant to the camera lens and switched them on and off in a predefined manner. For authentic eyes the sequence of detected reflections should match the sequence of stimulating near-infrared flashes. For non-eye or flat objects the reflections should either be missing, or placed in incorrect locations, making the detected sequence of reflections different from the original one. {\bf The method proposed in this paper is different from the above methods} in that it does not detect and use reflections from the cornea. Also, this approach of detecting corneal reflections may be fooled by contact lenses.

Another idea employing 3D features of the eye is based on detection of the {\it Purkinje reflections}, \ie specularities that occur at the outer and inner boundaries of the cornea, and the outer and inner boundaries of the lens. Lee \etal \cite{Lee_ICB_2005} follow this idea and apply a human eye model to calculate a theoretical positions of Purkinje reflections used later to verify the correctness of the observed specularities. {\bf The method proposed in this paper is different from the above method} since it does not detect and use Purkinje reflections in the PAD.  
Also, the higher-order Purkinje reflections can be difficult to detect and may still exist for a person wearing contact lenses.

An idea based on photometric stereo approach and illumination from different directions was proposed by Lee and Park \cite{Lee_IMA_2010}. They use the fact that the surface of a live iris is not perfectly flat and so will cast shadows when illuminated from different directions. In turn, a flat iris printout does not cast shadows. Properties of the surface estimated by photometric stereo method are then used to distinguish between a live iris and a flat printout. {\bf The method proposed in this paper is different from the above method} since it is designed for detection of textured contact lenses (not paper printouts) and it makes different assumptions on three-dimensional properties of the observed objects. In our method, and for image resolutions used in commercial systems, we assume that iris is more flat than the artifacts (textured contact lenses) we want to detect. This is an opposite assumption to the one made by Lee and Park, and the photometric stereo is applied in different ways in both methods.

An approach using a light stripe projected on the observed object was proposed by Connell \etal \cite{Connell_ASSP_2013}. This structured-light-based approach is able to assess the three-dimensional properties of the observed object, and thus provides a rough estimate whether the sensor is observing a live (approximately flat) iris, or a textured contact lens (more curved than authentic iris). However, this approach requires more elaborate sensor hardware than that used in current commercial iris sensors. Hughes and Bowyer \cite{Hughes_HICSS_2013} used stereo imaging to classify the observed texture as coming from a surface better approximated as planar (authentic iris) or spherical (textured contact lens). This approach requires that the sensor has an additional camera, and that the two cameras be separated well enough for the stereo to work well. {\bf The method proposed in this paper is different from all of the above methods} since it is does not use structured-light-based or stereo-vision image acquisition, and thus it does not require custom hardware (light-stripe projection or additional camera) to acquire images. And certainly, due to different acquisition techniques employed in our method, the image processing methodology is different than those presented in \cite{Connell_ASSP_2013} and \cite{Hughes_HICSS_2013}.

Our method requires nothing more than a pair of images that could be acquired from current commercial iris sensors in the course of normal image acquisition.

%-------------------------------------------------------------------------
\section{Method} % Disclosure}
\label{sec:Method}

\subsection{General Photometric Stereo Approach}

Photometric stereo is a computer vision method of estimating the surface normal vectors by observing an object under illuminations from different directions by a single fixed-position camera. Assuming we use $k$ point-wise illuminators that generate $k$ Lambertian reflections (\ie diverging almost equally in all directions) from each point of a surface with uniform albedo, we can use a linear model binding these quantities:

\begin{equation}
\mathbf{I} = \mathbf{L}\mathbf{\hat{n}} = \mathbf{L}c\mathbf{n}
\label{eqn:i}
\end{equation}

\noindent
where $\mathbf{I}$ is a vector of the observed $k$ intensities, $\mathbf{L}$ is a $3 \times k$ matrix of $k$ known light directions, $c$ represents a uniform albedo, and $\mathbf{n}$ is the surface unit normal vector to be estimated. This yields

\begin{equation}
\mathbf{\hat{n}} = 
\begin{cases}
    \mathbf{L}^{-1} \mathbf{I} & \text{if} \; k = 3 \\
    (\mathbf{L}^{T} \mathbf{L})^{-1}\mathbf{L}^{T}\mathbf{I} & \text{if} \; k \neq 3
\end{cases}
\label{eqn:nhat}
\end{equation}

\noindent
where $(\mathbf{L}^{T} \mathbf{L})^{-1}\mathbf{L}^{T}$ is  Moore-Penrose pseudoinverse of $\mathbf{L}$. Assuming the albedo $c$ is uniform for all  points, we can estimate the unit surface normals as

\begin{equation}
\mathbf{n} = \frac{\mathbf{\hat{n}}}{\|\mathbf{\hat{n}\|}}
\label{eqn:n}
\end{equation}

\noindent
where $\|x\|$ is the $\ell^2$ (Euclidean) norm of $x$. Since we need to find two unknowns in $\mathbf{\hat{n}}$, $k=2$ images taken under two different lighting conditions, are necessary to solve the equation (\ref{eqn:n}) and calculate $\mathbf{n}$. If three images are available, one may also recover albedo $c$ for each picture point. Having more than three images overdetermines the solution and allows to get more accurate estimations of both $\mathbf{n}$ and $c$.  However, our formulation of the PAD problem as a binary classification of the surface normals of the observed texture does not require more than two images.

The normal vectors $\mathbf{n}$ are estimated for each picture point $(x,y)$, so $\mathbf{I}_{x,y}$ and $\mathbf{n}_{x,y}$ should be used for the observed intensities and normal vector at point $(x,y)$, respectively. We skipped $(x,y)$ subscript in equations (\ref{eqn:i} - \ref{eqn:n}) for clarity.

\subsection{Application Context}

The photometric stereo, which is the core of the proposed PAD method, uses two near-infrared illuminators placed at different, yet known positions with respect to the lens, as already implemented in current commercial sensors. Thus this method can be immediately applied in current equipment without the need of hardware adaptations. Iris recognition sensors equipped with two not concentric illuminators automatically select their optimal configuration to generate a good quality iris image, compliant with ISO/IEC 19794-6 standard. This is important in particular when a subject wears glasses. Thus, in each presentation at least two iris pictures taken under illumination from different directions are available for these sensors, and can be used to check the presence of the textured contact lens according to the presented approach.

Extension of the presented concept to iris recognition sensors implementing more than two illuminators is straightforward, and would increase the accuracy of the estimation of 3D features. However, current commercial iris sensors commonly use just two illuminators.

\subsection{Detection of Textured Contact Lenses}

\begin{figure*}[!htb]
\centering
\begin{subfigure}[t]{0.51\textwidth}
\includegraphics[width=\textwidth]{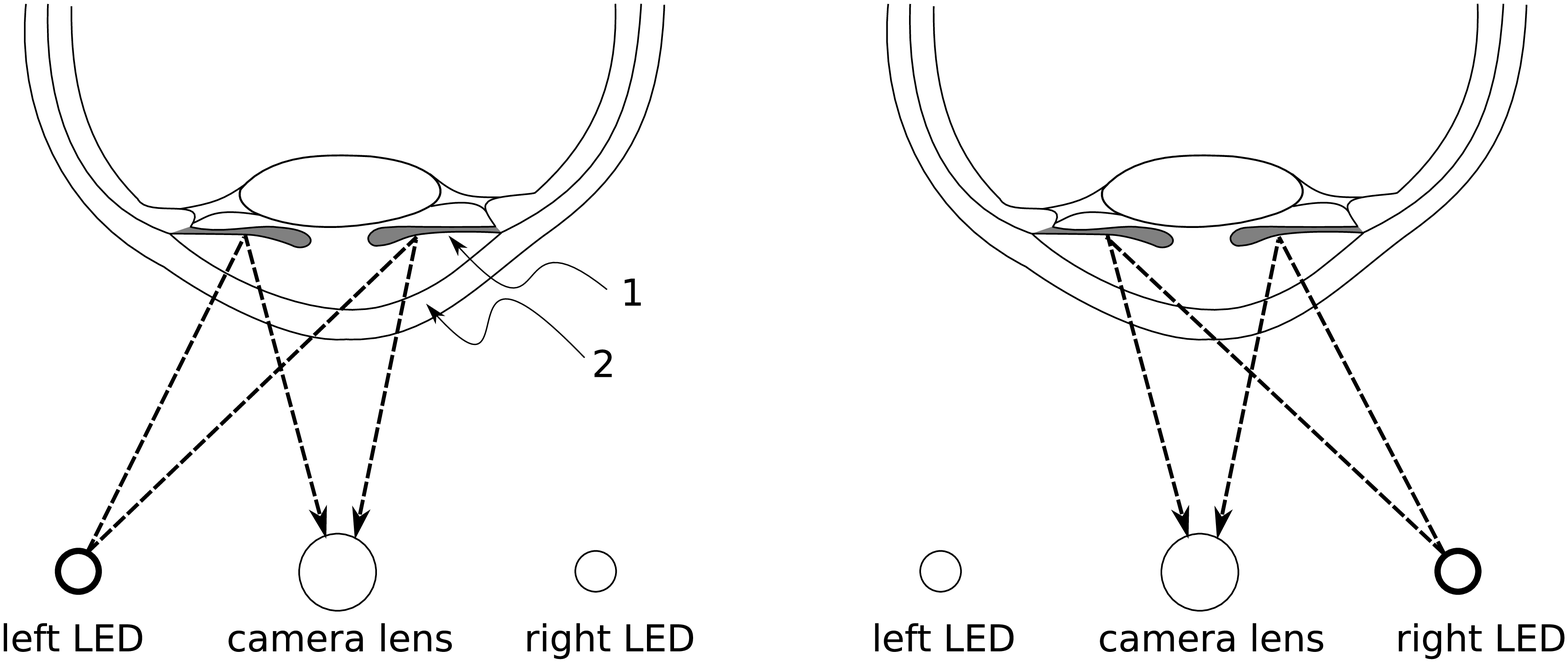}
\caption{~}
\end{subfigure}\hfill
\begin{subfigure}[t]{0.43\textwidth}
\hfill\frame{\includegraphics[width=0.48\textwidth]{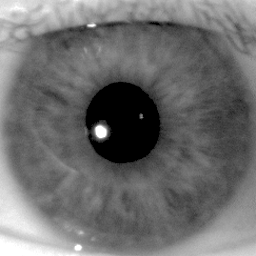}}\hskip1mm
\frame{\includegraphics[width=0.48\textwidth]{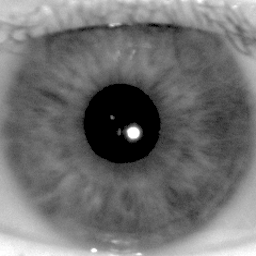}}
\caption{~}
\end{subfigure}
\caption{a) In case of observing an eye not wearing a texture contact lens, or wearing a transparent contact lens, the NIR light rays go through the cornea (2) and are reflected from the iris (1); b) The corresponding $I_{\mbox{\tiny left}}$ and $I_{\mbox{\tiny right}}$ iris images. Differences in shadows observed in these two pictures are small, which ends up with a reconstruction of a roughly planar surface.}
\label{fig:e1}
\end{figure*}

As stated before, the proposed presentation attack detection method is valid for arbitrary number of images taken under different lighting conditions. However, the presented experiments and results relate to the most popular setup in commercial iris sensors in which only two spatially separated illuminators are available, and thus $k=2$ iris images are acquired for this PAD, $I_{\mbox{\tiny left}}$ and $I_{\mbox{\tiny right}}$. Hence, we use two observed intensities in equation (\ref{eqn:i}) for each picture point $(x,y)$:

$$
\mathbf{I}_{x,y} = 
\begin{bmatrix}
 	I_{\mbox{\tiny left}}(x,y)\\
 	I_{\mbox{\tiny right}}(x,y)
 \end{bmatrix}
$$

Figure \ref{fig:e1}a illustrates a typical setup of two near-infrared illuminators placed equidistant to the camera lens. This allows to generate a pair of iris images shown in Fig. \ref{fig:e1}b. The external surface (visible to us) of the iris can be considered as a more Lambertian than specular surface. So, using either left or right illuminator produces very similar iris images. Certainly, the iris surface is not perfectly flat and this should manifest in different shadows visible in the left and right images. However, the resolution of commercial iris recognition sensors compliant to ISO/IEC 19794-6 is rather small (normally approx. 200 pixels across iris diameter are used, with 120 pixels being the standard requirement) when compared to the size of three-dimensional objects such as crypts, and hence the observed differences between $I_{\mbox{\tiny left}}$ and $I_{\mbox{\tiny right}}$ are small. The photometric stereo method will end up with estimation of normal vectors that should not differ too much from the average normal vector estimated for this object.

\begin{figure*}[!htb]
\centering
\begin{subfigure}[t]{0.51\textwidth}
\includegraphics[width=\textwidth]{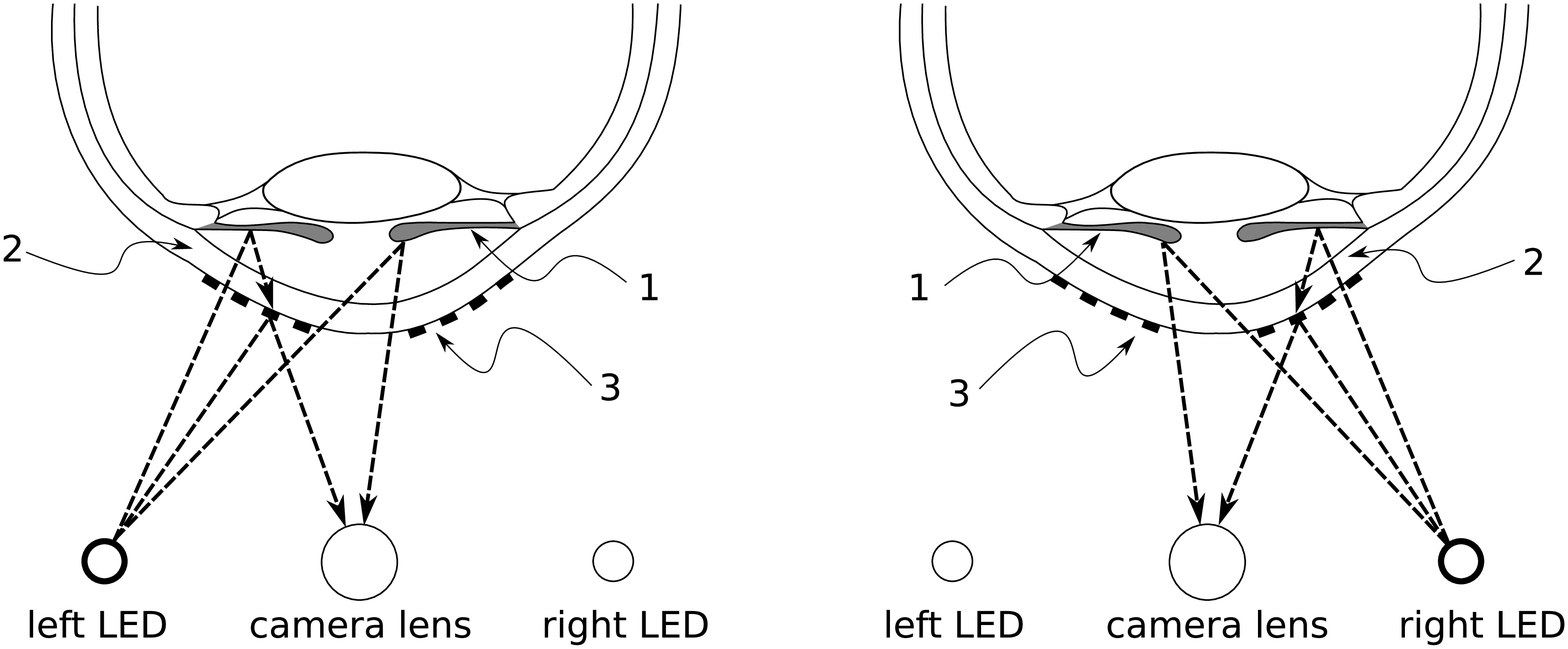}
\caption{~}
\end{subfigure}\hfill
\begin{subfigure}[t]{0.43\textwidth}
\hfill\frame{\includegraphics[width=0.48\textwidth]{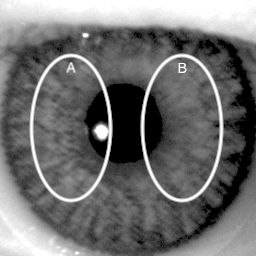}}\hskip1mm
\frame{\includegraphics[width=0.48\textwidth]{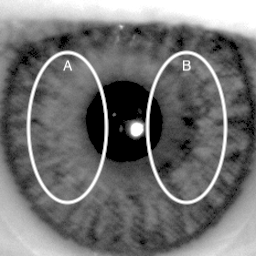}}
\caption{~}
\end{subfigure}
\caption{a) In case of observing an eye wearing a texture contact lens (3), the NIR light rays either go through the cornea (2) and, reflected from the iris, are registered by the camera, or they are reflected by the textured contact lens, or they are reflected from the iris but blocked by the opaque texture printed on the lens. b) The corresponding $I_{\mbox{\tiny left}}$ and $I_{\mbox{\tiny right}}$ iris images. Note that significant differences in generated shadows between the left- and right-illuminated iris with textured contact lens, especially in areas marked as A and B.}
\label{fig:e2}
\end{figure*}

Figure \ref{fig:e2} illustrates identical capture procedure (a) and the resulting images (b) when an eye wearing textured contact lens is photographed. One should note shadows made by the the partially opaque texture printed on the lens and observed in different places, depending on which illuminator was used to illuminate the object. Except for large shadows observed in regions marked as (A) and (B), we can also see differences how the printed texture generates image features under illumination at different angles. Consequently, for this object the photometric stereo will end up with highly variable normal vectors due to irregular and noisy surface that is being estimated.

\begin{figure}[!htb]
\centering
	\begin{subfigure}[t]{0.225\textwidth}
		\includegraphics[width=\textwidth]{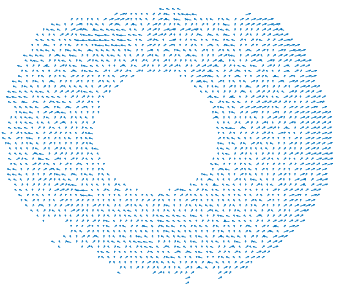}
		\caption{Live iris}
	\end{subfigure}\hfill
	\begin{subfigure}[t]{0.22\textwidth}
		\includegraphics[width=\textwidth]{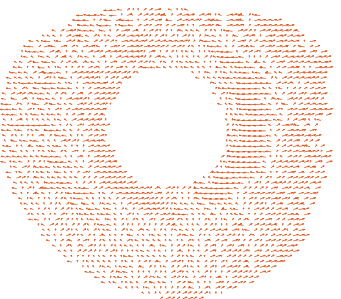}
		\caption{Iris with textured lens}
	\end{subfigure}
\caption{Quiver plots depicting estimated normal vectors for authentic iris (a) and iris with textured contact lens (b).}
\label{fig:quiver}
\end{figure}

The normal vectors estimated for the images shown in Figs. \ref{fig:e1}b and \ref{fig:e2}b are illustrated as quiver plots in Fig. \ref{fig:quiver}a and \ref{fig:quiver}b, respectively. Note a higher variability of the estimated normal vectors for an eye wearing a textured contact lens. Note also that we do not consider normal vectors outside the iris annulus, and for portions of the iris occluded by eyelids and eyelashes. This is accomplished by calculating the occlusion masks $m_{\mbox{\tiny left}}$ and $m_{\mbox{\tiny right}}$ corresponding to $I_{\mbox{\tiny left}}$ and $I_{\mbox{\tiny right}}$ iris images. For non-occluded iris pixels $m=1$, and for background pixels $m=0$.

\begin{figure}[!htb]
\centering
		\includegraphics[width=0.42\textwidth]{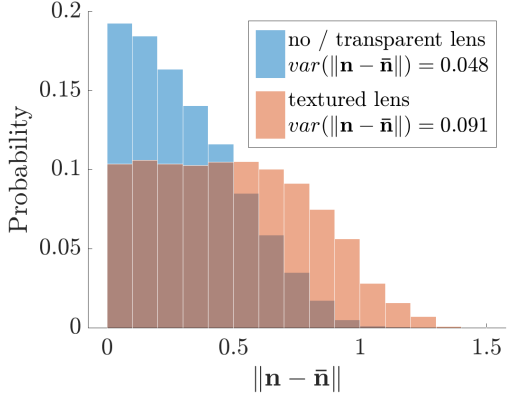}
\caption{Histograms illustrating how normal vectors $\mathbf{n}$ shown in Fig. \ref{fig:quiver} (a) and (b) differ from a mean normal vector $\mathbf{\bar{n}}$. As shown, variance of this difference is significantly larger when the textured contact lens is used.}
\label{fig:hist}
\end{figure}

So what distinguishes these two sets of normal vectors? Let $\mathbf{\bar{n}}$ be the average normal vector within the non-occluded iris area. Fig. \ref{fig:hist} presents distributions of vector norms between all normal vectors $\mathbf{n}$ and their averages $\mathbf{\bar{n}}$ calculated for iris with and without textured contact lens (shown in Figs. \ref{fig:e1}a and \ref{fig:e2}b). The Euclidean distances between normals and their average are smaller for an approximately flat iris when compared to an irregular object composed with an iris and a textured contact lens. Consequently, in this presentation attack detection method the variance of an Euclidean distance between the normals and their average, calculated in non-occluded iris area, is used as the PAD score:

\begin{equation}
\label{eqn:score}
q = var \| \mathbf{n}_{x,y} - \mathbf{\bar{n}} \|
\end{equation}

\noindent
where $\|x\|$ is the $\ell^2$ (Euclidean) norm of $x$, and

$$
\mathbf{\bar{n}} = \frac{1}{N}\sum_{x,y} \mathbf{n}_{x,y},
$$

\noindent
where $N$ is the number of non-occluded iris points, and
$$
\{(x,y): m_{\mbox{\tiny left}}(x,y) \cap m_{\mbox{\tiny right}}(x,y) = 1\}.
$$

We expect to observe a larger variance calculated by (\ref{eqn:score}) for irises wearing textured contact lenses than for irises not wearing textured contacts, or wearing transparent (clear) contacts.

\begin{figure}[!htb]
\centering
\includegraphics[width=0.46\textwidth]{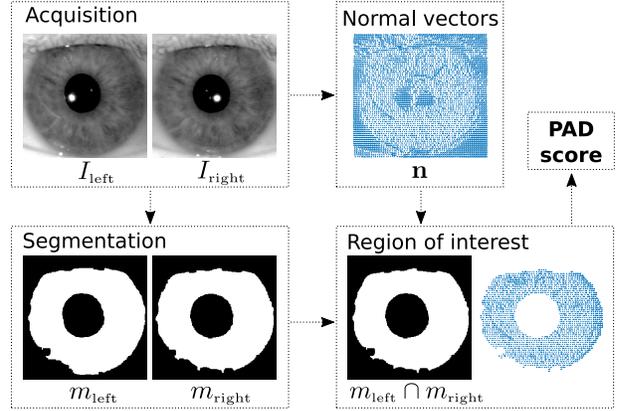}
\caption{Components of the proposed iris PAD method.}
\label{fig:normal}
\end{figure}

Figure \ref{fig:normal} presents all steps of the proposed iris PAD method, namely:

\begin{enumerate}
\item {\bf Iris image acquisition. } In this step, a pair of iris images, $I_{\mbox{\tiny left}}$ and $I_{\mbox{\tiny right}}$, is acquired. The $I_{\mbox{\tiny left}}$ image is acquired when the left illuminator is switched on, and the $I_{\mbox{\tiny right}}$ image is acquired when the right illuminator is switched on. Due to spontaneous oscillations in pupil size, the time between captures should be as small as possible.
\item {\bf Iris image segmentation. } In this step, occlusion masks, $m_{\mbox{\tiny left}}$ and $m_{\mbox{\tiny right}}$, are calculated, where $m=1$ denotes iris pixels, and $m=0$ denotes non-iris areas.
\item {\bf Estimation of normal vectors. } This step can be realized in parallel with the step 2, and ends up with a set of normal vectors estimated by photometric stereo method based on $I_{\mbox{\tiny left}}$ and $I_{\mbox{\tiny right}}$ images.
\item {\bf Setting the region of interest. } In this step normal vectors that are either outside of the iris annulus or correspond to eyelids and eyelashes are discarded from processing.
\item {\bf PAD score calculation and decision. } In this step the PAD score is calculated and a statistical test is applied to make a decision whether an observed iris is covered by a textured contact lens.
\end{enumerate}

%-------------------------------------------------------------------------
\section{Experiments and Results}
\label{sec:Results}

\begin{figure}[!htb]
\centering
\includegraphics[width=0.23\textwidth]{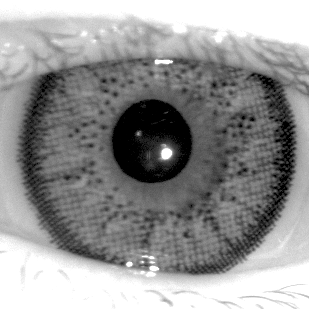}\hfill
\includegraphics[width=0.23\textwidth]{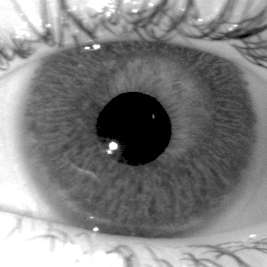}
\caption{Examples of textured contact lenses referred to as {\it regular}, \ie having a dot-like pattern ({\bf left}) and  {\it irregular}, \ie not having a dot-like pattern ({\bf right}).}
\label{fig:regular:irregular}
\end{figure}

\subsection{Database}

In this work, we used images from the Notre Dame Contact Lens Detection 2015 (NDCLD'15) dataset \cite{Doyle_Access_2015}. This is the only database known to us, which offers iris images (with and without contact lenses) of the same eyes captured shortly one after another with illumination coming from two different locations. Images acquired by the LG IrisAccess 4000 sensor are used to conduct the experiments, since this sensor was used to collect both images of authentic eyes and eyes with textured contact lenses. Additionally, this dataset contains images of eyes with clear (non-textured) contact lenses, used in this work to assess the impact of clear contact lenses on the method reliability. 

For this research we used iris image pairs, each with one image taken with left illuminator and the other with right illuminator, captured as close in time as possible. In total 5,796 iris images acquired from 119 subjects met this condition. This set was then divided into four subsets used in the experiments:

\begin{itemize}
    \item 1,800 images of irises wearing {\it regular} (with dot-like pattern) textured contact lenses, as shown in Fig. \ref{fig:regular:irregular}a,
    \item 864 images of irises wearing {\it irregular} (without dot-like pattern) textured contact lenses, as shown in Fig. \ref{fig:regular:irregular}b,
    \item 1,728 images of irises wearing clear contact lenses (without any visible pattern), and
    \item 1,404 images of authentic irises without any contact lenses (textured or clear).
\end{itemize}

Five different textured contact lens brands are represented in the NDCLD'15 dataset: Johnson \& Johnson, Ciba Vision, Cooper Vision, Clearlab and United Contact Lens. some subjects had images acquired multiple times wearing different types of textured contact lenses, and consequently the 2664 images (1,800 + 864) of irises with textured contact lenses represent 37 combinations of subject and contact lens brand.

\subsection{Experimental Scenarios}
\label{sec:splits}

% Three experiments with different training/testing splits are conducted: 1) images of regular textured contact lens (textured contact lens with clearly printed dot-like patterns) and proportionate number of randomly chosen images of authentic eyes are used in training, and images of irregular textured contact lens (textured contact lens that are colored but have no dot-like patterned) and the images of authentic eyes unseen in training are used in testing; 2) training set and testing set in 1) are swapped in this scenario to evaluate the capacity of generalization of the proposed method; 3) half of images of textured contact lens randomly chosen and half of images of authentic eyes are used in training, and the rest of the dataset are used in testing.

Three experiments with different compositions of training/testing subsets are conducted:

\begin{enumerate}[label=(\alph*)]

	\item images of {\it regular} textured contact lenses, \ie contact lenses with printed dot-like patterns, and proportionate number of randomly chosen images of authentic eyes (without clear contact lenses) are used in {\bf training}; images of eyes with {\it irregular} textured contact lenses, \ie those that do not have a dot-like pattern, and the images of authentic eyes (without clear contact lenses) {\bf unseen in training} are used in {\bf testing}; Fig. \ref{fig:regular:irregular} illustrates examples of regular and irregular contact lens patterns;

	\item training and testing sets in (a) are swapped in this scenario to evaluate the capacity of generalization of the proposed method; 

	\item regular and irregular texture contact lens patterns are mixed, then half of the images of textured contact lenses are randomly chosen for training along with images of authentic eyes (without clear contact lenses); the rest of the images of textured contact lenses, and images of eyes wearing clear contact lenses are chosen for testing. This scenario is repeated independently 10 times allowing for 10-fold cross validation (with replacement). The random selection of training images was continued until the training set has the same number of samples as the whole dataset. The aim of this experiment is to evaluate the influence of clear (not textured) contact lenses on the accuracy, assuming that no images of eyes wearing clear contact lenses were used in training. 

\end{enumerate}

\subsection{Error Metrics}
\label{sec:metrics}

We follow ISO/IEC 30107-3:2017 and use the following PAD error metrics: {\bf Attack Presentation Classification Error Rate (APCER)}, which is the proportion of {\it attack presentations} incorrectly classified as {\it bona fide presentations}, and {\bf Bona Fide Presentation Classification Error Rate (BPCER)}, which is the proportion of {\it bona fide presentations} incorrectly classified as {\it presentation attacks}. When {\it accuracy} is mentioned, it refers to the total number of correct classifications.

\subsection{Base Method}

\begin{figure*}[!htb]
\centering
	\begin{subfigure}[t]{0.46\textwidth}
		\includegraphics[width=\textwidth]{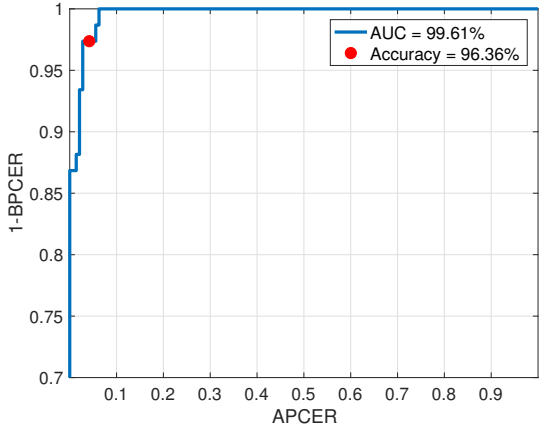}
		\caption{Trained on {\it regular}, tested on {\it irregular} contact lenses}
	\end{subfigure}\hfill
	\begin{subfigure}[t]{0.46\textwidth}
		\includegraphics[width=\textwidth]{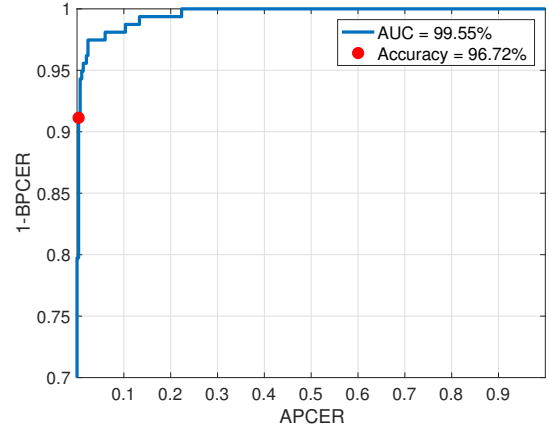}
		\caption{Trained on {\it irregular}, tested on {\it regular} contact lenses}
	\end{subfigure}
\caption{ROC curves achieved on the test set by the base method in the experiments (a) and (b), as defined in Sec. \ref{sec:splits}. Additionally, the APCER and BPCER achieved in testing for a threshold set on the training set are marked as a red dot.}
\label{fig:base_results}
\end{figure*}

% In the base method, we perform the proposed PAD method on the dataset using the entire non-occluded iris annulus. In each split, the PAD score that gives the Equal Error Rate (EER) for the training set is used as the classification threshold in the testing set. Results of training/testing splits a) and b) are shown in Fig. \ref{fig:base_results}. As shown in the figure, the proposed method gives satisfactory Receiver Operating Characteristic (ROC) curves for both split a) and b), with high area under the curve (AUC) values of 99.42\% and 99.6\%, respectively. A 10-fold cross validation of different splits c) is also conducted, whose result is shown in Fig. \ref{fig:notched_boxes}. In each fold of the validation, the training set consists of authentic iris images (with no clear contact lenses) and images of irises with textured contact lenses, both randomly chosen with replacement until the training set has the same number of samples as the whole dataset. Iris images unused in training are used for testing. The base method shows a stable and satisfactory performance, with an average accuracy of 97.50\%.

In the base method, we use the entire, non-occluded iris annulus to calculate the PAD score. Figure \ref{fig:base_results} presents the Receiver Operating Characteristic (ROC) curves for scenarios (a) and (b), as defined in Sec. \ref{sec:splits}, obtained on the test set. The areas under the curve (AUC), 99.61\% and 99.55\% for split (a) and (b), respectively, suggest that the effectiveness of the proposed method is high. Note that the method was trained and tested on different pattern types in scenarios (a) and (b), which means that this approach generalizes well to unknown contact lens patterns. The ROCs in Fig. \ref{fig:base_results} include also specific working points achieved when the threshold is set on the training data (based on the Equal Error Rate) and later used on the test data. This also suggests high generalization capabilities, due to a small difference between false positive and false negative rates in both (a) and (b) scenarios.

\subsection{Use of Weighted Local Areas}

\begin{figure}[!htb]
\centering
\includegraphics[width=0.46\textwidth]{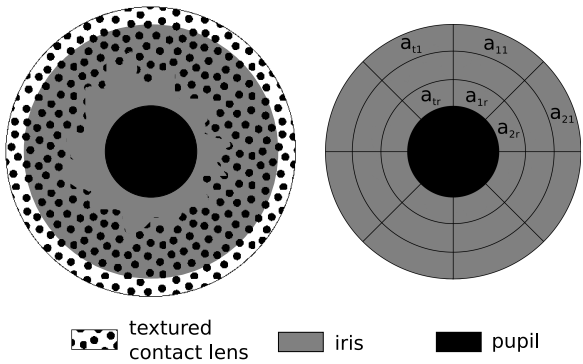}
\caption{{\bf Left:} Illustration of a typical placement of a textured contact lens. Contact lens pattern (dotted) often extends beyond the limbic (outer) iris boundary,  is irregular  in the central part and is not printed within the pupil area. {\bf Right:} The way how the iris annulus is divided to find the best local patches used in this PAD method.}
\label{fig:areas}
\end{figure}

\begin{figure*}[!htb]
\centering
	\begin{subfigure}[t]{0.46\textwidth}
		\includegraphics[width=\textwidth]{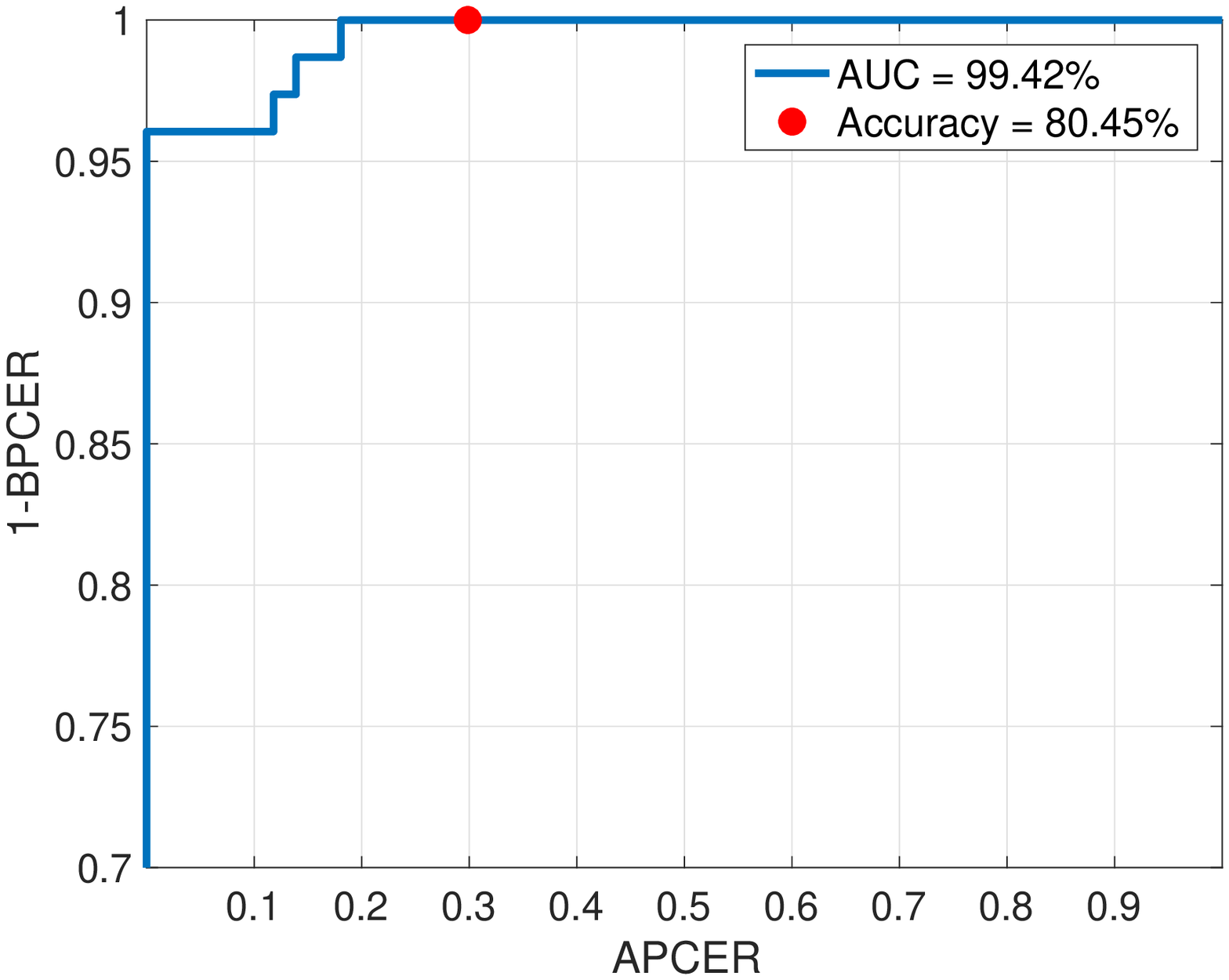}
		\caption{Trained on {\it regular}, tested on {\it irregular} contact lenses}
	\end{subfigure}\hfill
	\begin{subfigure}[t]{0.46\textwidth}
		\includegraphics[width=\textwidth]{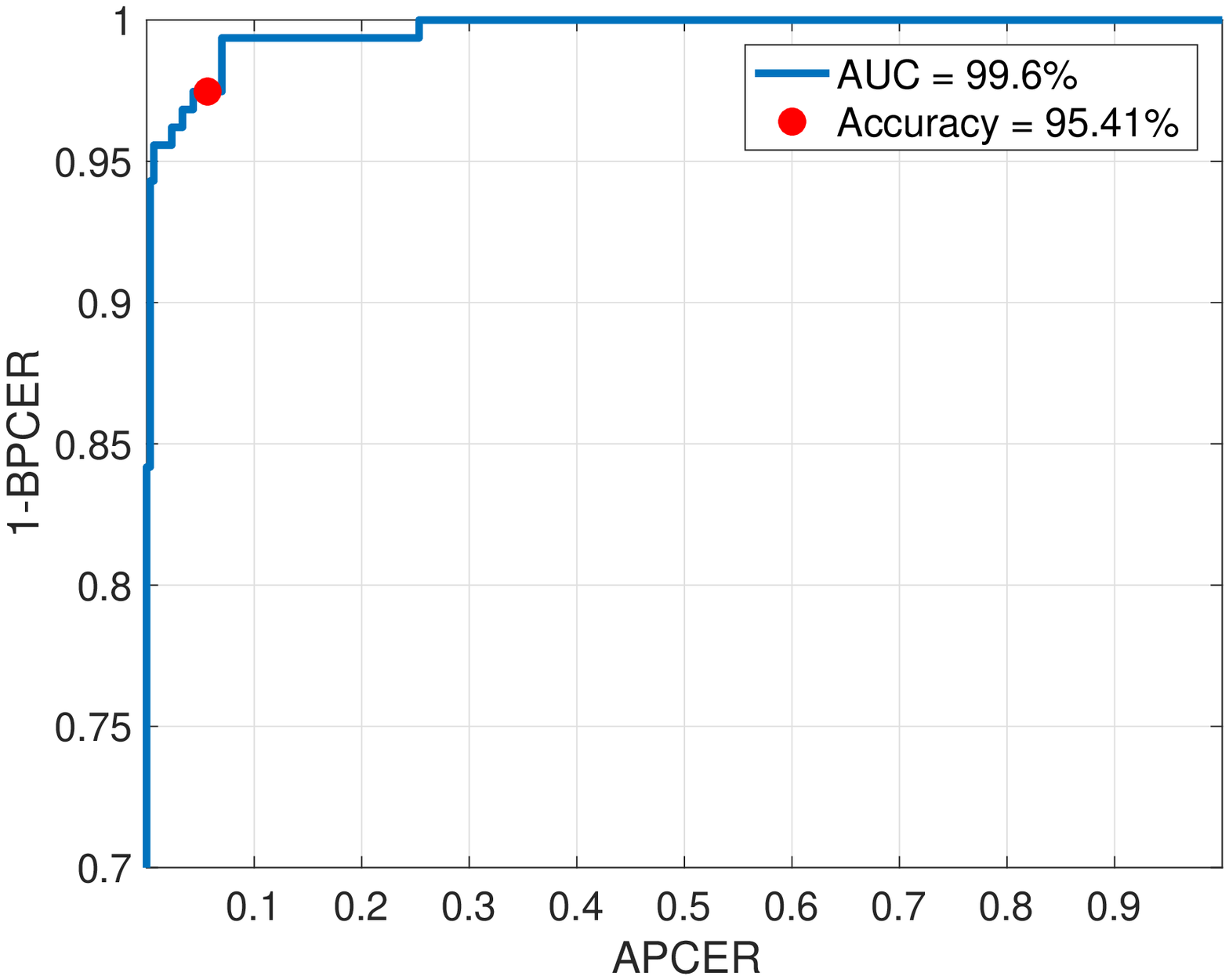}
		\caption{Trained on {\it irregular}, tested on {\it regular} contact lenses}
	\end{subfigure}
\caption{Same as in Fig. \ref{fig:base_results}, except that local areas were used in calculation of the PAD score.}
\label{fig:adv_results}
\end{figure*}

The estimated normal vectors may present nonuniform quality and usefulness for PAD across the non-occluded iris annulus. This is due to the following two reasons: (a) the segmentation may not be perfect, and the areas marked by the segmentation algorithm as a free-from-occlusions iris pattern may still contain eyelids, eyelashes or NIR reflections, (b) due to nonuniform construction of the textured contact lens (see Fig. \ref{fig:areas}, left) parts of the iris annulus close to the pupil will not often be covered by textured contact lens, and thus the PAD usability of these areas is limited.

We employ an iterative approach for searching the best local areas that, when combined together, offer an increased PAD accuracy when compared to using the entire non-occluded iris area. In each iteration, the annulus is segmented into $r$ radial and $t$ angular sections, ending up with $rt$ local sections $a_{i,j}$, for which the PAD scores can be calculated in the same way as for the entire non-occluded iris area, as depicted in Fig. \ref{fig:areas} (right). The usefulness of each local area is assessed by how well two distributions of PAD scores calculated for irises with and without textured contact lens are separated:

\begin{equation}
d'={\frac {\mu _{\mbox{\scriptsize authentic}}-\mu _{\mbox{\scriptsize contact}}}{\sqrt {0.5(\sigma _{\mbox{\scriptsize authentic}}^{2}+\sigma _{\mbox{\scriptsize contact}}^{2})}}}
\label{eqn:dprime}
\end{equation}

where $\mu _{\mbox{\scriptsize authentic}}$ and $\mu _{\mbox{\scriptsize contact}}$ represent the mean PAD scores of authentic irises and contact lenses, respectively, and $\sigma _{\mbox{\scriptsize authentic}}$ and $\sigma _{\mbox{\scriptsize contact}}$ denote the standard deviation of the PAD scores of authentic irises and contact lenses, respectively. Therefore, a significance map of the iris annulus is obtained, and it is used in the search for the best iris mask. We experimented with various numbers for $r$ and $t$, with $r\in\langle4,5\rangle$ and $t\in\langle10,15\rangle$ ending up with the best results.

Then, $rt$ steps of the mask selection are performed. In the $p$-th iteration, $p$ local areas with the highest $d'$ are included in the calculation of the PAD score:

\begin{equation}
q_{\mbox{\scriptsize weighted}} = {\frac{\sum_{x,y}d'_{x,y}(l_{x,y}-l_{\mbox{\scriptsize w}})^{2}}{\sum_{x,y}d'_{x,y}}}
    \label{eqn:weightedscore}
\end{equation}

\noindent
where $d'_{x,y}$ is the $d'$ of the local area $a_{i,j}$ point ${x,y}$ is in, and

$$
l_{\mbox{\scriptsize w}}={\frac{\sum_{x,y}d'_{x,y}l_{x,y}}{\sum_{x,y}d'_{x,y}}}, \quad l_{x,y}=\| \mathbf{n}_{x,y} - \mathbf{\bar{n}} \|^{2}
$$

After the PAD score calculation in each iteration, the $d'$ of the two distributions of PAD scores are again calculated for irises with and without textured contact lenses. The highest $d'$ defines the final selection of local areas $a_{i,j}$ used in testing. This iterative procedure is performed only to find the best local regions for a given database, and there is no need to repeat it for each iris image pair being verified.

Results of the experiments (a) and (b) using the proposed weighted local area selection are presented in Fig. \ref{fig:adv_results}. This variant of the method again generates satisfactory ROC curves and high AUC values of 99.42\% and 99.6\%, respectively. However, this variant presents lower generalization capabilities, since the threshold set on the train set allowed to achieve 80.45\% of accuracy when irises with regular textured contact lenses are used for training, and irregular patterns are seen in testing. When regular and irregular patterns are represented in both the training and testing sets (experiment (c), as defined in Sec. \ref{sec:splits}), the variant in which local weighted areas are used presents slightly better average accuracy in 10-fold cross-validation experiment (98.38\%) than the base variant (97.50\%), as depicted in Fig. \ref{fig:notched_boxes}. 

\subsection{Do Clear Contact Lenses Impact Accuracy?}

\begin{figure}[!htb]
\centering
\includegraphics[width=0.41\textwidth]{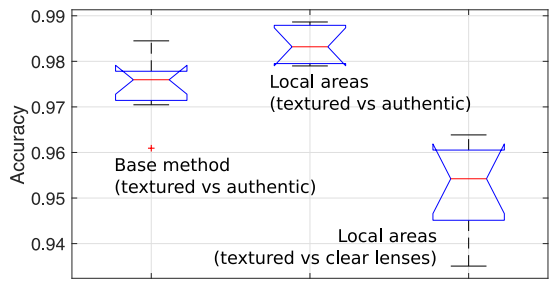}
\caption{Notched box plots for 10-fold cross validation of a base method, a variant employing weighted local areas tested on irises with textured contact lenses and clear eyes, and the same variant tested on irises with textured and clear contact lenses.}
\label{fig:notched_boxes}
\end{figure}

What happens when the method is trained on image pairs acquired from eyes without any contact lenses, and then tested on subjects wearing transparent contact lenses? In the iris PAD, only textured contact lenses should be detected as a presentation attack, while an eye wearing clear contact lens should be recognized as an authentic eye. In this subsection we present experimental results to explore the impact of clear contact lenses on the method performance. A 10-fold cross validation is carried out to evaluate the ability of the proposed method to correctly classify irises wearing clear contact lenses. In each fold of the validation, the training set is created as defined in the experiment (c) in Sec. \ref{sec:splits}, while the testing set consisted of randomly chosen images of irises wearing clear contact lenses and images of irises wearing textured contact lenses unused in training. Fig. \ref{fig:notched_boxes} shows that the average accuracy of 95.24\% is slightly lower than 98.38\% achieved when the eyes are free from any contact lenses. This is expected, since even clear contact lenses have a boundary between the active lens area and the transparent carrier, which also may generate shadows, increasing our PAD score. At the same time, we can conclude that the impact of clear contact lenses on this method is small, and the obtained accuracy about 95\% is acceptable. 

%-------------------------------------------------------------------------
\section{Conclusions}

This paper presents the first photometric-stereo-based iris presentation attack detection method designed to detect textured contact lenses. There are two important properties of the proposed solution:

\begin{itemize}
    \item {\bf good generalization capabilities:} since the proposed algorithm is not trained for any specific pattern, and uses a simple observation that textured contact lenses will produce larger shadows than authentic iris when illuminated at different angles, this PAD is agnostic to contact lens brand or specific pattern printed on the lens; this is in general not true for image filtering-based feature extractors (including deep-learning-based solutions), which are known to perform worse in the open-set testing scenario \cite{Pinto_DLB_2018,Yambay_IJCB_2017};
    
    \item {\bf immediate application to commercial iris sensors:} since this method uses the hardware elements (illuminators, camera) already present in commercial sensors, there are no required hardware updates needed to add this method as one of the PAD techniques; the requirement to collect two iris images illuminated from two different angles is easily met, since iris sensors typically do this in each acquisition to minimize the impact of specular reflections, especially from glasses. 
\end{itemize}

The method was developed on a database of approx. 5,800 iris images (2,900 image pairs) acquired from approx. 100 subjects. This method is able to correctly classify more than 98\% samples if the textured contact lens brand is known, and offers an accuracy of above 95\% in open-set scenario when the contact lens brand is unknown in testing. Due to lack of other photometric-stereo-based iris PAD algorithms, and lack of other publicly available databases suitable for research on photometric-stereo-based iris PAD, the comparison with state-of-the-art methods was not possible. The source codes of the method are offered along with this paper to other researchers. 

%  However, 

{\small
\bibliographystyle{ieee}
\bibliography{refs}
}

\end{document}